\makeatletter
\def\input@path{{./}{ICLR2027/}{../ICLR2027/}}
\makeatother
\documentclass{article}

\usepackage{iclr2027_conference,times}

\usepackage{amsmath,amsfonts,bm}

\def\eqref#1{equation~\ref{#1}}
\def\1{\bm{1}}

\DeclareMathAlphabet{\mathsfit}{\encodingdefault}{\sfdefault}{m}{sl}
\SetMathAlphabet{\mathsfit}{bold}{\encodingdefault}{\sfdefault}{bx}{n}

\usepackage[table]{xcolor}
\usepackage{graphicx}
\graphicspath{{Figures/}{ICLR2027/Figures/}{../ICLR2027/Figures/}}
\usepackage{booktabs}
\usepackage{multirow}
\usepackage{algorithm}
\usepackage{algorithmic}
\usepackage{float}
\usepackage{amsmath}
\usepackage{hyperref}
\usepackage{url}
\usepackage{wrapfig}
\usepackage{threeparttable}

\newcommand{\relaxkvg}{RelaxKV\smash{\ensuremath{^{\mathrm{G}}}}}
\title{RelaxKV: Recomputation Guided by the Query with Sparse Context Attention for Efficient KV Cache Reuse}

\iclrfinalcopy

\author{
\textbf{Ruoling Qi}$^{1,2,*}$ \quad
\textbf{Yirui Liu}$^{2,*,\dagger}$ \quad
\textbf{Xuaner Wu}$^{2}$ \quad
\textbf{Yuxin Jin}$^{2}$ \quad
\textbf{Jian Chen}$^{3}$ \\
\textbf{Jiayu Qin}$^{3}$ \quad
\textbf{Yin Chen}$^{2}$ \quad
\textbf{Jiawei Shao}$^{2,\dagger}$ \\[1.5mm]
$^{1}$Shanghai Jiao Tong University \\
$^{2}$Institute of Artificial Intelligence, China Telecom (TeleAI)\\
$^{3}$State University of New York at Buffalo \\[1mm]
$^{*}$Equal contribution
\qquad
$^{\dagger}$Corresponding authors
}

\begin{document}

\maketitle

\begin{abstract}
Cross-request KV caching reduces the prefill cost of Retrieval-Augmented Generation (RAG), but conventional prefix caching severely limits cache reuse across requests. Position-Independent Caching (PIC) removes this constraint by reusing independent chunks, but their KV states miss cross-chunk interactions. Existing methods selectively recompute token states to recover these missing interactions, but primarily allocate the recomputation budget to selecting which states to recompute, while fixing the recomputation context to the full causal prefix. We introduce \textbf{RelaxKV}, which formulates selective cache repair as a joint allocation problem over repair targets and recomputation context. Guided by the user query, RelaxKV identifies layer-specific repair targets and restricts their recomputation to a query-relevant context, reducing attention computation. Across four decoder models, RelaxKV at a 15\% anchor ratio improves aggregate LongBench performance over ProphetKV on all models. On Qwen3-14B, RelaxKV provides a stronger quality–TTFT trade-off than ProphetKV across a 5\%–30\% anchor-ratio sweep, and achieves the best selective results on RULER-MV and LV-Eval at 16K and 32K context lengths. Controlled ablations further demonstrate the importance of recomputation context selection. 
Our code will be released upon acceptance.
\end{abstract}

% ======================================
\section{Introduction}

Retrieval-Augmented Generation (RAG) often retrieves overlapping document chunks across requests, making cross-request Key-Value (KV) cache reuse an effective way to reduce prefill cost \citep{lewis2020rag,gao2023ragsurvey}. Conventional prefix caching, however, only reuses KV states when requests share the same prefix, limiting reuse across requests. Position-Independent Caching (PIC) removes this restriction by precomputing chunks independently, but the resulting KV states miss cross-chunk interactions when the chunks are composed. Full recomputation restores these interactions but loses the efficiency benefit of cache reuse, while selective cache repair recomputes only part of the context to recover them at lower online cost.

Existing selective repair methods mainly differ in how they choose repair targets. EPIC repairs fixed boundary states, CacheBlend and KVShare select targets based on cache or hidden-state deviations, and ProphetKV uses query relevance \citep{hu2025epic,yao2025cacheblend,yang2025kvshare,wang2026prophetkv}. Despite these different selection strategies, they share the same recomputation pattern: once a token is selected, it still attends to its full causal prefix during recomputation. Their recomputation budget therefore controls which token states are repaired, but not how much context each repair reads. As a result, even a small target set can still incur substantial attention computation, exposing a second design dimension: \emph{recomputation context selection}.

We introduce RelaxKV, which formulates selective cache repair as a joint allocation problem over \emph{repair targets} and \emph{recomputation context}. RelaxKV first uses query attention to select repair anchors independently across layers. Deeper selections require the corresponding token states to be recomputed through the preceding layers, with each token remaining active only through its highest selected layer. This yields layer-specific active repair targets, while the union of all layer-wise anchors defines the KV positions available as attention context. At each layer, an active target attends only to the causally preceding positions in this union: active positions contribute newly recomputed KV states, while the remaining positions reuse their cached KV states. In this way, RelaxKV jointly controls which token states are recomputed and which KV entries they attend to, reducing both state and attention computation.

Experiments across four decoder models and multiple long-context benchmarks demonstrate the effectiveness of RelaxKV. At a 15\% anchor ratio, RelaxKV improves aggregate LongBench performance over ProphetKV on all four models. On Qwen3-14B, sweeping the anchor ratio from 5\% to 30\% shows a stronger quality--TTFT trade-off than ProphetKV on RULER-MV. RelaxKV also achieves the best selective performance on both RULER-MV and LV-Eval at 16K and 32K context lengths. To isolate the role of recomputation context, we further evaluate RelaxKV$^{G}$, a matched-target control that retains ProphetKV's repair targets while changing only the recomputation context. Together with context-construction ablations, these results show that recomputation context selection is an important factor in the quality--latency trade-off.

Our main contributions are summarized as follows:
\begin{itemize}
    \item We formulate selective cache repair as a joint allocation problem over two dimensions: \emph{repair target selection}, governing state recomputation, and \emph{recomputation context selection}, governing attention-context exposure.

    \item We introduce RelaxKV, which derives layer-specific repair targets and a shared query-relevant recomputation context from layer-wise query attention.

    \item Experiments demonstrate competitive quality–latency trade-offs across models, with particularly strong gains on long-context benchmarks.
\end{itemize}

% ====================================
\section{Background and Motivation}

\subsection{Approximate Prefix KV Reuse}

For a transformer with only a decoder \citep{vaswani2017attention} and $L$
layers, a full prefill of a token sequence $x_{1:T}$ produces a KV cache
indexed by layer,
$\mathcal{K}^{\mathrm{full}}=\{(K_l,V_l)\}_{l=0}^{L-1}$.  In a reusable cache
setting, the context is divided into chunks and each chunk is prefetched
independently.  Concatenating these chunk caches gives a cheap cache
$\mathcal{K}^{\mathrm{naive}}$, but generally
$\mathcal{K}^{\mathrm{naive}}\ne\mathcal{K}^{\mathrm{full}}$ because the
hidden state of a token was computed without all preceding chunks.

\paragraph{Positional consistency.}
Rotary position embedding (RoPE) rotates queries and keys according to their
sequence positions, making their inner product depend on the relative
displacement between tokens \citep{su2024roformer}.  To separate this
positional effect from contextual staleness across chunks, our evaluation
prefetches each reusable chunk using its final global position IDs rather
than positions that restart at the chunk boundary.  If a cached chunk is
later placed at a different offset, the RoPE rotation identity can realign
its cached keys before repair \citep{chen2026kvpacket}.  This positional
operation is orthogonal to RelaxKV: even a positionally aligned cache remains
approximate because each chunk was encoded without the hidden states of
preceding chunks.  RelaxKV addresses this missing context from preceding chunks rather
than RoPE phase misalignment.

We use the following three cache states throughout the paper. \emph{Full reuse} directly uses the precomputed chunk caches without repair. \emph{Full recompute} performs a dense prefill over the complete context. \emph{Selective repair} starts from the reusable cache, selectively recomputes token states, and updates their corresponding KV entries. RelaxKV operates within the selective repair setting; it does not modify cache construction or model weights.

\subsection{Query-Aware Cache Repair}

\begin{wrapfigure}{r}{0.5\textwidth}
    \vspace{-12pt}  
    \centering    
    \includegraphics[width=\linewidth]{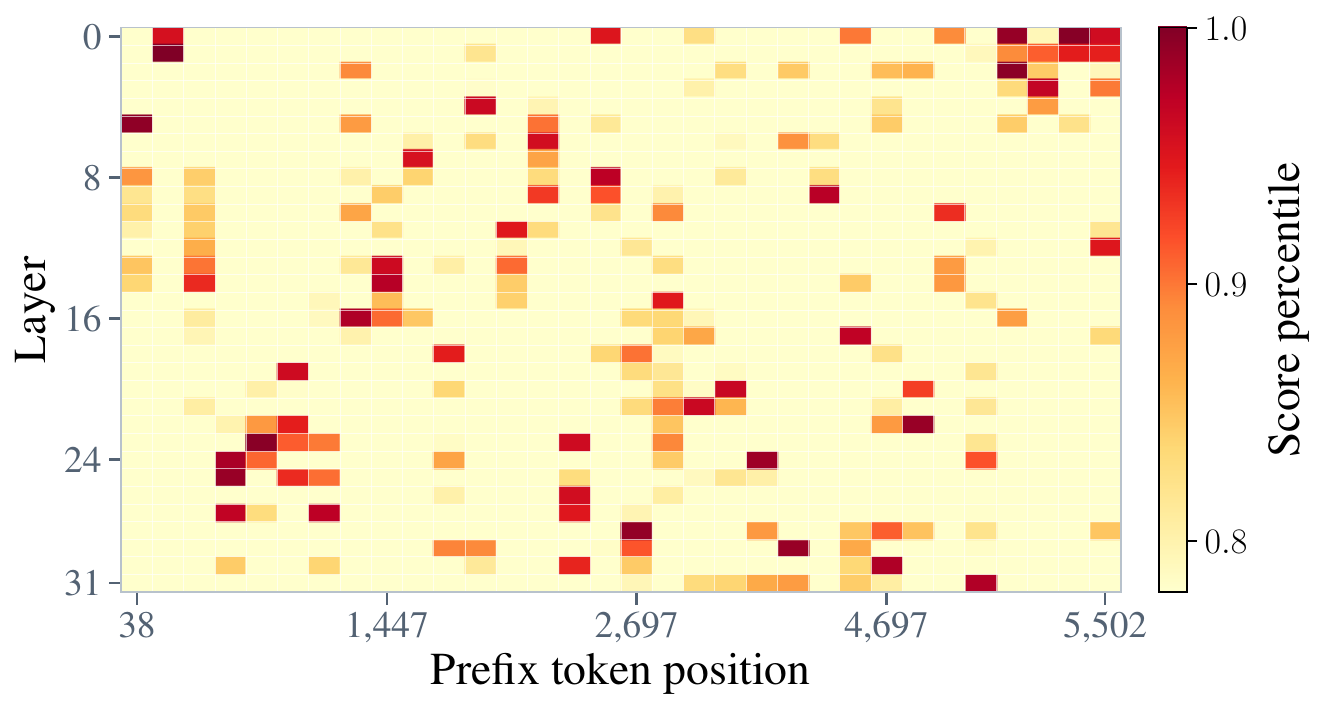}
    \caption{Layer-wise query-attention scores over prefix tokens. High-attention score tokens positions shift across layers.}
    \label{fig:layer_specific}
\end{wrapfigure}

Let $q$ denote the user query appended after the cached context. Running $q$ against $\mathcal{K}^{\mathrm{naive}}$ produces query-to-context attention at every layer. As illustrated in Figure~\ref{fig:layer_specific}, highly attended token positions can vary substantially across layers. We therefore pool the attention weights at each layer into a score $a_{l,i}$ for context position $i$, and use these layer-specific scores to guide repair. More generally, selective repair involves two decisions: which token states to recompute, and which cached KV states those tokens can attend to during recomputation. Repair target selection decides which states are recomputed, whereas recomputation context selection decides which KV states are used during their recomputation. Existing methods mainly optimize repair target selection while treating the full causal prefix as a fixed recomputation context.

This distinction motivates the design of RelaxKV. A token selected at a particular layer is a \emph{repair anchor}. If the same token is selected at a deeper layer, its state must be recomputed through the preceding layers required to reach that layer, yielding the layer-specific \emph{active repair targets}. Separately, \emph{recomputation context tokens} contribute the keys and values available to these active targets during attention. Distinguishing repair anchors, active repair targets, and recomputation context tokens makes explicit the two dimensions of selective repair: which attention rows are recomputed and which cached KV columns they can access.

% ====================================
\section{Method}
% ----------------------------
\subsection{Overview}

RelaxKV repairs a reusable chunk KV cache for a given query by jointly selecting repair targets and their recomputation context. Given the naive cache $\mathcal{K}^{\mathrm{naive}}$, a query $q$, and an anchor ratio $r$, RelaxKV first uses layer-wise query attention to select repair anchors independently at each layer. A token selected at a deeper layer is recomputed through the preceding layers required to reach its highest selected layer, producing layer-specific active repair targets. The union of the layer-wise anchors forms a shared recomputation context.

\begin{figure*}[t]
    \centering
    \includegraphics[width=\textwidth]{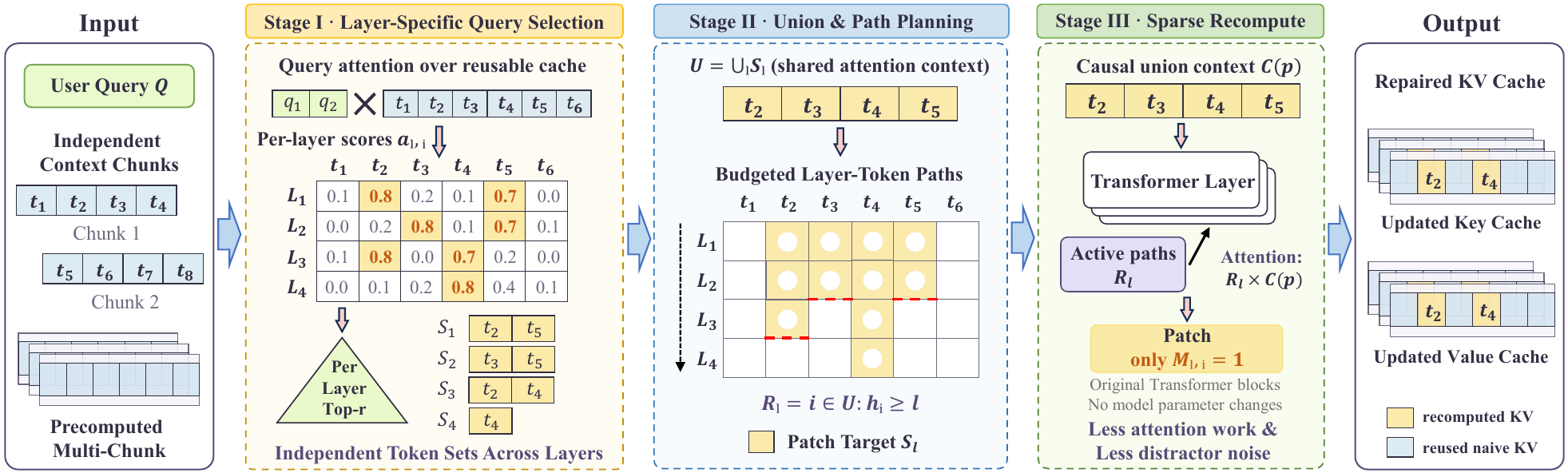}
    \caption{\textbf{Overview of RelaxKV.} Layer-wise query attention selects repair anchors, from which RelaxKV derives layer-specific active repair targets and a shared recomputation context. At each layer, active targets attend only to the causally valid positions in this context and update their corresponding KV entries.}
    \label{fig:relaxkv-overview}
\end{figure*}

At each layer, only the active repair targets execute the Transformer block. Each target attends to the causally valid subset of the shared context and updates its corresponding KV entry. Thus, repair target selection determines which token states are recomputed, while recomputation context selection determines which cached KV states they can attend to. Figure~\ref{fig:relaxkv-overview} summarizes the workflow.

% ----------------------------------
\subsection{Layer-wise Query Attention and Repair Targets}

We run the query with $\mathcal{K}^{\mathrm{naive}}$ and collect its attention to the cached context. For layer $l$ and context position $i$, we average over heads and query positions and normalize within each layer:
\[
    a_{l,i}=\operatorname{MinMax}
    \left(\frac{1}{HQ}\sum_{h=1}^{H}\sum_{t=1}^{Q}
    A_l[h,t,i]\right).
\]
The score $a_{l,i}$ is used for selection; recomputation still executes the original Transformer blocks.

At each layer, RelaxKV independently selects the $k=\max(1,\operatorname{round}(rT_c))$ context positions with the highest scores, where $T_c$ is the number of context tokens and $r$ is the anchor ratio. We denote the selected set at layer $l$ by $S_l$ and refer to its elements as \emph{repair anchors}. Independent layer-wise selection allows the repair targets to adapt to changes in token relevance across model depth rather than forcing all layers to share a single global target set.

A token selected at a deeper layer must be recomputed through the preceding layers required to reach that layer. For each token $i$ appearing in at least one anchor set, we define its highest selected layer as
\[
\begin{aligned}
    h_i &= \max\{l:i\in S_l\},\\
    U   &= \bigcup_l S_l,\\
    R_l &= \{i\in U:h_i\ge l\}.
\end{aligned}
\]
Here, $U$ is the union of all layer-wise anchors, and $R_l$ is the set of active repair targets at layer $l$. A token remains active through its highest selected layer $h_i$ and becomes inactive thereafter. Thus, $\{R_l\}_{l=0}^{L-1}$ closes the layer-wise anchor selections under the dependencies required for recomputation. The nominal anchor ratio $r$ controls the number of selected anchors, while the resulting number of layer-token executions can be larger because deeper selections require recomputation through preceding layers.

% -----------------------
\subsection{Recomputation Context Selection}

RelaxKV constructs the recomputation context from the shared anchor union $U$. At layer $l$, only active repair targets in $R_l$ execute the Transformer block. For an active target at its original sequence position $p$, we define the causal recomputation context as
\[
    C_l(p)=U\cap\{j:j\le p\}.
\]
Each target therefore attends only to the causally valid positions in $U$, rather than its full causal prefix. We gather the cached keys and values at these positions and replace entries in \(R_l\cap C_l(p)\) with freshly recomputed KV states before computing attention. The resulting hidden states then pass through the original output projection, residual connections, normalization, and MLP. Tokens with $h_i=l$ complete their repair at this layer and leave the active set.

All active repair targets retain their original global position IDs when applying RoPE. Restricting attention to $C_l(p)$ therefore changes only which KV positions are available during recomputation, without altering the positional coordinates or relative offsets of the retained token pairs.

\begin{figure}[t]
    \centering
    \includegraphics[width=\columnwidth]{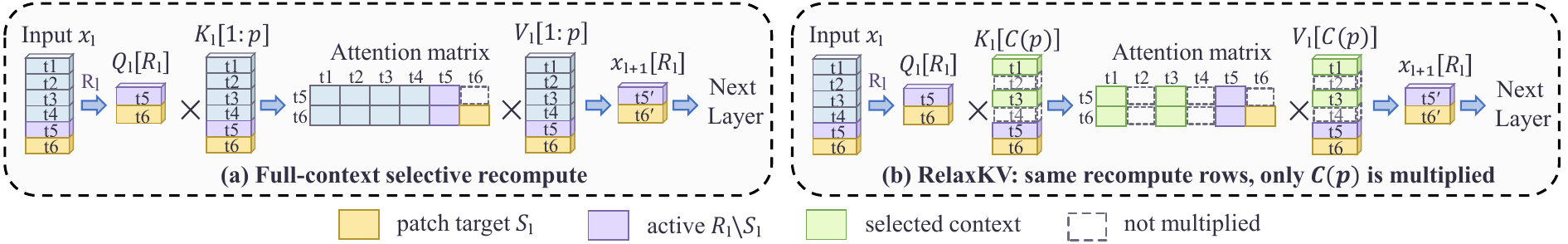}
    \caption{\textbf{Full-context versus RelaxKV recomputation.} Both execute the same active repair targets $R_l$. For a target at position $p$, full-context recomputation attends to the complete causal prefix, whereas RelaxKV attends only to $C_l(p)=U\cap\{j:j\le p\}$. Dashed entries are excluded from the recomputation context.}
    \label{fig:recompute-context}
\end{figure}

Figure~\ref{fig:recompute-context} contrasts full-context recomputation with RelaxKV under the same active repair targets $R_l$. The two settings differ only in the KV positions that each target can attend to: full-context recomputation uses the complete causal prefix, whereas RelaxKV uses only $C_l(p)$.

% -----------------------------------
\subsection{Cache Update Semantics}

Recomputing the active targets produces fresh KV states for every $i\in R_l$. These states overwrite the corresponding entries in the reusable cache, while all other entries retain their naive values. The update set is therefore determined directly by $R_l$ and requires no additional selection step.

During recomputation, however, the attention context can contain both fresh and cached KV states. For each position $j\in C_l(p)$ at layer $l$, RelaxKV uses
\[
 (\widetilde K_{l,j},\widetilde V_{l,j})=
 \begin{cases}
 (K^{\mathrm{new}}_{l,j},V^{\mathrm{new}}_{l,j}), & j\in R_l\cap C_l(p),\\
 (K^{\mathrm{naive}}_{l,j},V^{\mathrm{naive}}_{l,j}), & j\in C_l(p)\setminus R_l.
 \end{cases}
\]
Thus, an active target attends to freshly recomputed KV states for positions that are also active at the current layer, while reusing cached KV states for the remaining positions in its recomputation context. Once a token reaches its highest selected layer and leaves the active target set, its KV states at deeper layers are not recomputed and retain their naive values. The token may nevertheless remain in the shared context $U$ and contribute these cached KV states during later recomputation. This mixed fresh--cached KV context is part of RelaxKV's approximation and is used consistently in all comparisons.

% ====================================================
\section{Evaluation}
\label{sec:evaluation}

Our evaluation addresses four questions. (1) Does RelaxKV preserve task quality across models and long-context workloads? (2) Does its effectiveness persist as context length increases? (3) When repair targets are fixed, how does recomputation context selection affect quality and cost? (4) Does jointly allocating repair targets and recomputation context improve the end-to-end quality--TTFT trade-off across repair budgets?

\subsection{Experimental Setup}

\paragraph{Models.}
We evaluate four decoder models spanning 3B to 14B parameters: Llama-3.1-8B-Instruct~\citep{grattafiori2024llama3}, Qwen3-14B~\citep{yang2025qwen3}, Phi-4-14B~\citep{abdin2024phi4}, and Llama-3.2-3B-Instruct~\citep{meta2024llama32}. Llama and Phi use instruction-tuned checkpoints, while Qwen3 is evaluated with its default thinking mode enabled.

\paragraph{Tasks and metrics.}
We evaluate controlled long-context retrieval with RULER-MV~\citep{hsieh2024ruler} and natural long-context tasks with LongBench~\citep{bai2024longbench} and LV-Eval~\citep{yuan2024lveval}, using the standard metric for each benchmark. For LongBench, we select seven QA, multi-hop QA, and passage-retrieval tasks that naturally match the retrieval-based reusable-cache setting, including MuSiQue~\citep{trivedi2022musique}, while the length stress tests evaluate contexts from 4K to 32K tokens.

\paragraph{Implementation and hardware.}
Experiments are conducted on NVIDIA H100 80GB HBM3 GPUs using PyTorch 2.8.0, CUDA 12.9, and Transformers 4.57.1, with BF16 model weights and greedy decoding. All selective repair methods run on a single GPU. Only the 16K and 32K full-recomputation reference runs use two GPUs with model parallelism and SDPA.

\paragraph{Baselines and protocol.}
We evaluate cache repair after retrieval, with the same retrieved chunks shared across all methods and retrieval latency excluded from the measured request time. Within each comparison, directly comparable methods use identical prompts, tokenization, model weights, 512-token chunks, and evaluation examples. We compare RelaxKV with full recomputation, full reuse by directly concatenating the precomputed chunk caches, and four selective repair methods: CacheBlend~\citep{yao2025cacheblend}, EPIC~\citep{hu2025epic}, KVShare~\citep{yang2025kvshare}, and ProphetKV~\citep{wang2026prophetkv}. For Llama-3.2-3B, we additionally report KVLink using its publicly released checkpoint and blocked-attention inference~\citep{yang2025kvlink}.

\paragraph{Control with Fixed ProphetKV Targets.}
Let $P$ denote ProphetKV's global repair target set, which is shared across layers. We use \relaxkvg{} as a matched-target diagnostic control: it retains ProphetKV's repair targets $P$ and cache update mask, but replaces the full causal prefix with RelaxKV's recomputation context. Specifically, a target at position $p$ attends to $U\cap\{j:j\le p\}$, where $U$ is constructed from the layer-wise query-attention anchors at the same nominal ratio. Thus, \relaxkvg{} keeps repair target selection and cache updates fixed while changing only the recomputation context. We report it as a diagnostic control rather than a separate method.

In the main comparisons, selective baselines and \relaxkvg{} use a 20\% repair ratio, whereas RelaxKV uses an anchor ratio of $r=0.15$. These nominal ratios are not directly equivalent to the realized computation of RelaxKV: selecting anchors at deeper layers requires recomputation through preceding layers, while their union determines the size of the recomputation context. We therefore separately report the realized context-set and active-target ratios in Sec.~\ref{sec:4.5} and use measured TTFT to characterize actual online cost.

% ------------------------------------
\subsection{Accuracy on Standard Tasks}

Table~\ref{tab:main-results} reports task-level and aggregate LongBench results. Directly comparable selective baselines and \relaxkvg{} use a nominal 20\% repair ratio, while RelaxKV uses a 15\% anchor ratio. Under these settings, RelaxKV improves the aggregate score over ProphetKV on all four models while maintaining performance close to the full-recomputation reference. Under matched ProphetKV targets, \relaxkvg{} achieves aggregate scores within a small margin of ProphetKV across all models, showing that changing the recomputation context alone can preserve comparable task quality. At the task level, no single repair strategy dominates every dataset, but RelaxKV remains consistently competitive across multihop QA and passage-retrieval tasks.

\newcommand{\ourscell}[1]{#1}
\newcommand{\mainmodelblank}{\cellcolor{white}\strut}
\newcommand{\oursavg}[1]{\cellcolor{gray!21}#1}
\newcommand{\mainmethodrow}{\rowcolor{gray!10}}

\begin{table*}[t]
\centering
\caption{\textbf{LongBench results.} Full Recompute serves as the dense reference. Directly comparable selective baselines and \relaxkvg{} use a nominal 20\% repair ratio, while RelaxKV uses a fixed 15\% anchor ratio. Bold and underline denote the best and second-best selective results, respectively.}
\label{tab:main-results}

\begingroup
\scriptsize
\renewcommand{\arraystretch}{1.02}
\newcolumntype{T}{w{c}{62pt}}
\newcolumntype{D}{w{c}{22pt}}
\newcolumntype{A}{>{\columncolor{gray!12}}w{c}{24.5pt}}
\setlength{\tabcolsep}{0.35pt}

\resizebox{\textwidth}{!}{%
\begin{minipage}{1.27\textwidth}

\begin{minipage}[t]{0.49\linewidth}
\centering
\begin{tabular}{T|*{7}{D}A}
\toprule
\multicolumn{9}{c}{\textbf{Qwen3-14B}} \\
\cmidrule(lr){1-9}
\textbf{Method} & \textbf{WQA} & \textbf{TQA} & \textbf{HQA} & \textbf{NQA} & \textbf{MQue} & \textbf{PR-en} & \textbf{PR-zh} & \textbf{LB Avg} \\
\midrule
\textcolor{gray}{Full Recompute} & \textcolor{gray}{8.50} & \textcolor{gray}{88.41} & \textcolor{gray}{54.95} & \textcolor{gray}{24.15} & \textcolor{gray}{25.57} & \textcolor{gray}{68.00} & \textcolor{gray}{100.00} & \textcolor{gray}{52.80} \\
\textcolor{gray}{Full Reuse} & \textcolor{gray}{7.22} & \textcolor{gray}{66.60} & \textcolor{gray}{28.01} & \textcolor{gray}{12.45} & \textcolor{gray}{5.47} & \textcolor{gray}{24.00} & \textcolor{gray}{72.14} & \textcolor{gray}{30.84} \\
CacheBlend & 8.46 & 85.11 & 42.57 & 21.57 & 20.57 & 66.00 & 77.00 & 45.90 \\
EPIC & 8.52 & \textbf{87.23} & 47.63 & 23.31 & 21.11 & 53.00 & 88.00 & 46.97 \\
KVShare & 7.46 & 86.47 & 33.46 & 13.64 & 7.03 & 36.50 & 69.04 & 36.23 \\
ProphetKV & \textbf{8.64} & 86.78 & \underline{52.24} & \underline{24.20} & \underline{25.92} & \textbf{74.00} & \textbf{100.00} & \underline{53.11} \\
\relaxkvg{}$_{.20}$ & \underline{8.55} & 86.70 & 50.94 & \textbf{24.61} & 23.89 & \textbf{74.00} & \textbf{100.00} & 52.67 \\
\mainmethodrow
\ourscell{\textbf{RelaxKV}$_{.15}$} & \ourscell{8.32} & \ourscell{\underline{86.84}} & \ourscell{\textbf{52.41}} & \ourscell{23.88} & \ourscell{\textbf{26.52}} & \ourscell{\textbf{74.00}} & \ourscell{\textbf{100.00}} & \oursavg{\textbf{53.14}} \\
\bottomrule
\end{tabular}
\end{minipage}
\hfill
\begin{minipage}[t]{0.49\linewidth}
\centering
\begin{tabular}{T|*{7}{D}A}
\toprule
\multicolumn{9}{c}{\textbf{Llama-3.1-8B}} \\
\cmidrule(lr){1-9}
\textbf{Method} & \textbf{WQA} & \textbf{TQA} & \textbf{HQA} & \textbf{NQA} & \textbf{MQue} & \textbf{PR-en} & \textbf{PR-zh} & \textbf{LB Avg} \\
\midrule
\textcolor{gray}{Full Recompute} & \textcolor{gray}{47.23} & \textcolor{gray}{88.01} & \textcolor{gray}{49.10} & \textcolor{gray}{24.73} & \textcolor{gray}{25.86} & \textcolor{gray}{69.00} & \textcolor{gray}{85.71} & \textcolor{gray}{55.66} \\
\textcolor{gray}{Full Reuse} & \textcolor{gray}{27.28} & \textcolor{gray}{83.68} & \textcolor{gray}{33.28} & \textcolor{gray}{15.85} & \textcolor{gray}{9.65} & \textcolor{gray}{11.00} & \textcolor{gray}{11.25} & \textcolor{gray}{27.43} \\
CacheBlend & 37.24 & 84.93 & 45.68 & 21.03 & 20.15 & 33.00 & 37.88 & 39.99 \\
EPIC & 35.43 & 87.91 & \textbf{51.76} & 22.34 & 19.31 & 44.00 & 61.92 & 46.10 \\
KVShare & 29.35 & 88.73 & 36.01 & 21.47 & 18.56 & 19.50 & 24.00 & 33.95 \\
ProphetKV & 45.21 & 90.30 & \underline{49.04} & \underline{23.13} & \underline{25.88} & 73.50 & \underline{93.33} & 57.20 \\
\relaxkvg{}$_{.20}$ & \underline{45.86} & \textbf{90.80} & 48.62 & \textbf{25.30} & 25.40 & \underline{74.50} & \textbf{94.43} & \textbf{57.84} \\
\mainmethodrow
\ourscell{\textbf{RelaxKV}$_{.15}$} & \ourscell{\textbf{48.24}} & \ourscell{\underline{90.46}} & \ourscell{48.32} & \ourscell{22.96} & \ourscell{\textbf{25.93}} & \ourscell{\textbf{76.00}} & \ourscell{91.60} & \oursavg{\underline{57.64}} \\
\bottomrule
\end{tabular}
\end{minipage}

\par\smallskip

\begin{minipage}[t]{0.49\linewidth}
\centering
\begin{tabular}{T|*{7}{D}A}
\toprule
\multicolumn{9}{c}{\textbf{Phi-4-14B}} \\
\cmidrule(lr){1-9}
\textbf{Method} & \textbf{WQA} & \textbf{TQA} & \textbf{HQA} & \textbf{NQA} & \textbf{MQue} & \textbf{PR-en} & \textbf{PR-zh} & \textbf{LB Avg} \\
\midrule
\textcolor{gray}{Full Recompute} & \textcolor{gray}{22.02} & \textcolor{gray}{87.74} & \textcolor{gray}{38.41} & \textcolor{gray}{13.83} & \textcolor{gray}{9.11} & \textcolor{gray}{65.25} & \textcolor{gray}{97.33} & \textcolor{gray}{47.67} \\
\textcolor{gray}{Full Reuse} & \textcolor{gray}{12.51} & \textcolor{gray}{81.13} & \textcolor{gray}{19.88} & \textcolor{gray}{7.88} & \textcolor{gray}{4.91} & \textcolor{gray}{18.00} & \textcolor{gray}{23.00} & \textcolor{gray}{23.90} \\
CacheBlend & 15.47 & 82.99 & 28.84 & 8.23 & 4.72 & 50.00 & 37.10 & 32.48 \\
EPIC & 18.81 & 85.34 & 31.21 & 8.12 & 5.44 & 49.00 & 55.00 & 36.13 \\
KVShare & 14.79 & 84.82 & 32.01 & 8.58 & 6.13 & 42.50 & 47.08 & 33.70 \\
ProphetKV & \textbf{20.48} & \underline{87.01} & \underline{36.19} & \textbf{8.90} & \underline{8.38} & \underline{67.83} & 89.90 & \underline{45.53} \\
\relaxkvg{}$_{.20}$ & 18.97 & 86.19 & 33.39 & \underline{8.69} & 8.14 & 66.64 & \underline{90.24} & 44.61 \\
\mainmethodrow
\ourscell{\textbf{RelaxKV}$_{.15}$} & \ourscell{\underline{19.86}} & \ourscell{\textbf{87.59}} & \ourscell{\textbf{38.40}} & \ourscell{8.21} & \ourscell{\textbf{8.51}} & \ourscell{\textbf{69.06}} & \ourscell{\textbf{92.28}} & \oursavg{\textbf{46.27}} \\
\bottomrule
\end{tabular}
\end{minipage}
\hfill
\begin{minipage}[t]{0.49\linewidth}
\centering
\begin{tabular}{T|*{7}{D}A}
\toprule
\multicolumn{9}{c}{\textbf{Llama-3.2-3B}} \\
\cmidrule(lr){1-9}
\textbf{Method} & \textbf{WQA} & \textbf{TQA} & \textbf{HQA} & \textbf{NQA} & \textbf{MQue} & \textbf{PR-en} & \textbf{PR-zh} & \textbf{LB Avg} \\
\midrule
\textcolor{gray}{Full Recompute} & \textcolor{gray}{29.49} & \textcolor{gray}{82.47} & \textcolor{gray}{48.06} & \textcolor{gray}{22.10} & \textcolor{gray}{18.57} & \textcolor{gray}{67.00} & \textcolor{gray}{40.00} & \textcolor{gray}{43.96} \\
\textcolor{gray}{Full Reuse} & \textcolor{gray}{14.84} & \textcolor{gray}{57.00} & \textcolor{gray}{19.66} & \textcolor{gray}{5.15} & \textcolor{gray}{5.91} & \textcolor{gray}{4.00} & \textcolor{gray}{7.18} & \textcolor{gray}{16.25} \\
CacheBlend & 18.84 & 57.89 & 20.72 & 5.95 & 5.07 & 3.00 & 7.45 & 16.99 \\
EPIC & 21.94 & 80.97 & 40.22 & 18.79 & 14.55 & 22.00 & 9.00 & 29.64 \\
KVShare & 20.63 & 77.59 & 25.52 & 12.99 & 13.37 & 5.50 & 5.53 & 23.02 \\
KVLink$^\dagger$ & 17.69 & \textbf{86.32} & 44.24 & 19.32 & 6.92 & 26.00 & 5.66 & 29.45 \\
ProphetKV & \underline{29.13} & \underline{85.33} & 43.82 & 19.90 & 16.66 & \textbf{72.50} & 24.00 & 41.62 \\
\relaxkvg{}$_{.20}$ & 28.74 & 84.83 & \underline{45.82} & \underline{20.07} & \underline{17.17} & 68.50 & \underline{30.00} & \underline{42.16} \\
\mainmethodrow
\ourscell{\textbf{RelaxKV}$_{.15}$} & \ourscell{\textbf{30.17}} & \ourscell{84.41} & \ourscell{\textbf{46.71}} & \ourscell{\textbf{20.22}} & \ourscell{\textbf{17.78}} & \ourscell{\underline{70.00}} & \ourscell{\textbf{39.00}} & \oursavg{\textbf{44.04}} \\
\bottomrule
\end{tabular}
\end{minipage}

\end{minipage}%
}

\endgroup

\vspace{1pt}

\parbox{\textwidth}{
\scriptsize
\raggedright
$\dagger$ KVLink uses the publicly released Llama3B-KVLink5 checkpoint.
}

\end{table*}

% ---------------------------------
\subsection{Context Length Scaling}

Table~\ref{tab:longcontext-ruler} evaluates Qwen3-14B on RULER-MV from 4K to 32K context lengths and on LV-Eval at 16K and 32K. Selective baselines and \relaxkvg{} use a 20\% repair ratio, while RelaxKV uses its standard 15\% anchor ratio. On RULER-MV, Full Reuse degrades rapidly as the context grows, while selective repair remains substantially more robust. At 32K, RelaxKV achieves 95.00 accuracy, compared with 91.17 for ProphetKV and 92.00 for \relaxkvg{}, and remains close to the Full Recompute reference.

LV-Eval provides a complementary setting with multiple documents and natural distractors. RelaxKV achieves the best selective performance at both 16K and 32K, reaching 28.71 and 25.40 token F1, compared with 24.78 and 19.32 for ProphetKV. Together, these results reveal an additional benefit of recomputation context selection beyond controlling computation: as the context becomes longer and contains more irrelevant information, restricting repair to query-relevant context can better preserve quality under a limited repair budget. This pattern is consistent with an implicit filtering, or denoising, effect during recomputation.

\begin{table}[t]
\centering
\caption{\textbf{Long-context results on Qwen3-14B.}  RULER-MV reports accuracy and LV-Eval reports token F1.}
\label{tab:longcontext-ruler}
\begingroup
\scriptsize
\setlength{\tabcolsep}{5.5pt}
\begin{tabular}{l*{4}{c}|*{2}{c}}
\toprule
\multirow{2}{*}{\textbf{Method}}
& \multicolumn{4}{c|}{\textbf{RULER-MV} $\uparrow$}
& \multicolumn{2}{c}{\textbf{LV-Eval} $\uparrow$} \\
\cmidrule(lr){2-5}\cmidrule(lr){6-7}
& \textbf{4K} & \textbf{8K} & \textbf{16K} & \textbf{32K}
& \textbf{16K} & \textbf{32K} \\
\midrule
\textcolor{gray}{Full Recompute}
& \textcolor{gray}{100.00} & \textcolor{gray}{100.00}
& \textcolor{gray}{96.67} & \textcolor{gray}{95.83}
& \textcolor{gray}{32.37} & \textcolor{gray}{23.27} \\
\textcolor{gray}{Full Reuse}
& \textcolor{gray}{52.00} & \textcolor{gray}{41.17}
& \textcolor{gray}{38.67} & \textcolor{gray}{13.83}
& \textcolor{gray}{0.64} & \textcolor{gray}{0.29} \\
\midrule
CacheBlend & 69.33 & 55.50 & 41.50 & 20.33 & 1.66 & 0.74 \\
EPIC & 57.17 & 49.00 & 52.17 & 71.33 & 14.37 & 10.11 \\
KVShare & 53.67 & 47.17 & 44.33 & 24.00 & 0.98 & 0.38 \\
ProphetKV & \underline{96.67} & \textbf{98.83} & 92.83 & 91.17
& \underline{24.78} & \underline{19.32} \\
\relaxkvg{}$_{.20}$ & 96.17 & 97.83 & \underline{93.67} & \underline{92.00}
& 22.13 & 17.30 \\
\rowcolor{gray!10}
\textbf{RelaxKV}$_{.15}$
& \textbf{98.67} & \underline{98.67} & \textbf{95.33} & \textbf{95.00}
& \textbf{28.71} & \textbf{25.40} \\
\bottomrule
\end{tabular}
\endgroup
\end{table}

% ---------------------------------
\subsection{Ablation Study of Recomputation Context Construction}

Table~\ref{tab:recompute-context-ablation} isolates the effect of recomputation context construction while fixing the repair schedule. All variants use the same 15\% layer-wise anchor budget, identical anchors $S_l$, and the same active repair targets $R_l$. They also use the same nominal context-set size before causal truncation. Let $U_l$ denote the context set used at layer $l$; for a repair target at position $p$, the resulting causal context is $C_l(p)=U_l\cap\{j:j\le p\}.$ The variants differ only in how $U_l$ is constructed. \emph{Global Sparse} selects a single context set shared across layers using scores averaged over depth, \emph{Layerwise Sparse} selects an independent context set at each layer. RelaxKV uses the union of the layer-wise repair anchors.

Matching the context-set size controls the nominal number of retained KV positions, but does not guarantee identical attention work. Because causal truncation depends on the target position $p$, the effective edge count $\mathcal{E}_{\mathrm{unmasked}}=\sum_l\sum_{p\in R_l}|C_l(p)|$ can differ across constructions even when their pre-truncation context sizes match. We therefore treat this experiment as a controlled ablation of context construction rather than a strict equal-work comparison, and report measured recomputation latency alongside quality.

The results show that context construction matters even when the repair targets and nominal context budget are fixed. The three variants have similar recomputation latency, yet their task quality can differ substantially, indicating that recomputation context selection is not determined by sparsity alone: which KV positions are retained is itself an important design choice.

\begin{table*}[t]
\centering
\caption{Effect of recomputation context construction on Qwen3-14B at a
15\% anchor ratio. All variants share the same repair anchors,
dependency-closed active repair targets, and pre-truncation context-set size
$|U|$; only context construction varies. Panel (a) defines the variants, and
panel (b) reports task quality / mean recomputation latency (s).}
\label{tab:recompute-context-ablation}
\begingroup
\footnotesize
\renewcommand{\arraystretch}{1.14}
\setlength{\tabcolsep}{2.4pt}
\resizebox{\textwidth}{!}{%
\begin{tabular}[t]{@{}c@{\hspace{10pt}}c@{}}
\begin{tabular}[t]{@{}lccc@{}}
\toprule
\textbf{Variant} & \textbf{Context Source} & \textbf{Scope} & \textbf{Set size} \\
\midrule
Global Sparse & Across-Layer Mean & Shared & $|U|$ \\
Layerwise Sparse & Per-Layer Score & At Each Layer & $|U|$ \\
\rowcolor{gray!10}
\textbf{RelaxKV}$_{.15}$ & Anchor union $U$ & Shared & $|U|$ \\
\bottomrule
\end{tabular}
&
\begin{tabular}[t]{@{}lcccc@{}}
\toprule
\textbf{Variant} & \textbf{MuSiQue} & \textbf{RULER 32K}
& \textbf{LV-Eval 16K} & \textbf{LV-Eval 32K} \\
\midrule
Global Sparse & 34.81 / 5.97 & \textbf{95.00} / 31.87
& 27.32 / 25.44 & 25.35 / 78.99 \\
Layerwise Sparse & 31.78 / 5.89 & 85.00 / 31.86
& 18.18 / 25.48 & 21.76 / 79.02 \\
\rowcolor{gray!10}
\textbf{RelaxKV}$_{.15}$ & \textbf{38.08} / 5.84 & \textbf{95.00} / 31.72
& \textbf{28.71} / 25.50 & \textbf{25.40} / 78.39 \\
\bottomrule
\end{tabular}
\\[7pt]
\textbf{(a) Context Set Components}
& \textbf{(b) Quality / recomputation latency (s)}
\end{tabular}}
\endgroup
\end{table*}

% --------------------------------
\subsection{Budget Accounting and Online Serving Latency}
\label{sec:4.5}

\begin{figure*}[t]
    \centering
    \includegraphics[width=0.9\textwidth]{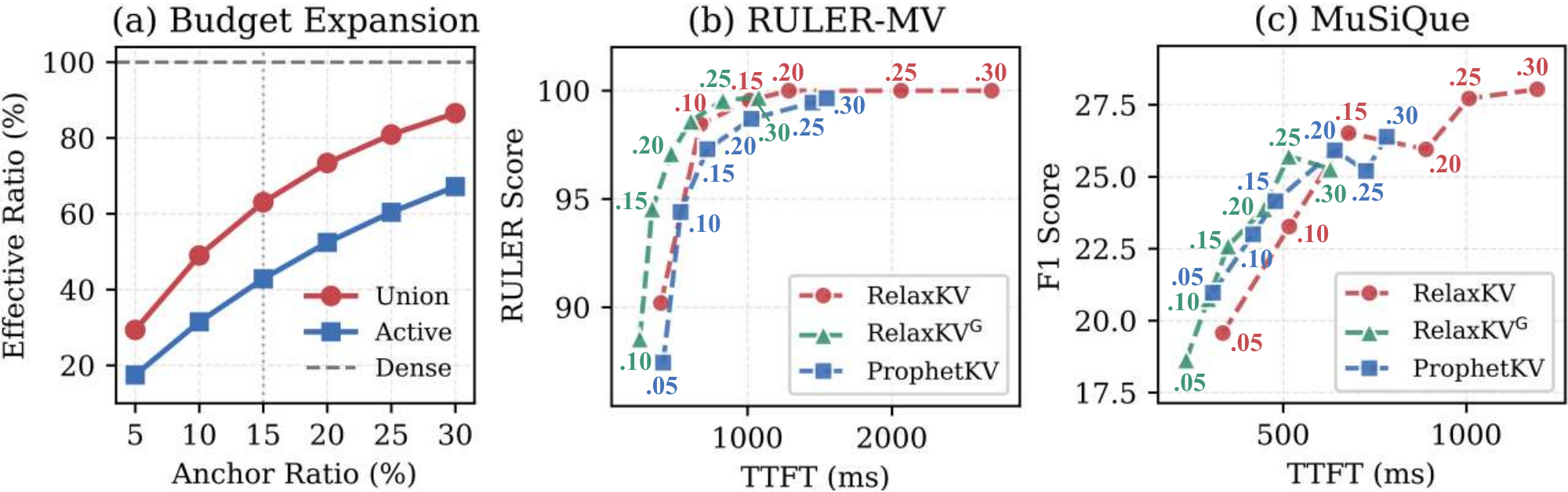}
    \caption{
    Qwen3-14B budget accounting and quality--TTFT trade-offs.
    (a) A layer-wise anchor ratio induces a shared-context union ratio and a
    dependency-closed active-target ratio. (b--c) Quality--latency trade-off on Qwen3-14B. We sweep
    nominal recomputation ratios from 5\% to 30\%; labels denote the ratios.
    RelaxKV reaches higher RULER-MV accuracy at lower TTFT.}
    \label{fig:quality-latency}
\end{figure*}

Figure~\ref{fig:quality-latency} relates RelaxKV's nominal anchor budget to its realized computation and end-to-end quality--TTFT trade-off on Qwen3-14B. Panel~(a) reports the shared-context union ratio and the dependency-closed active-target ratio as the layer-wise anchor ratio varies. Because anchors selected at deeper layers require recomputation through preceding layers, the nominal anchor ratio does not directly equal the fraction of executed layer-token states. At \(r=0.15\), for Qwen3-14B on the 8K RULER-MV workload with 512-token chunks, the shared anchor union covers 63.05\% of context positions, while the dependency-closed active targets account for 42.85\% of layer-token states, leaving 57.15\% unrecomputed. Panels~(b--c) report the corresponding quality and TTFT on RULER-MV and MuSiQue. As the anchor ratio increases, RelaxKV trades additional recomputation for higher task quality, producing a favorable quality--TTFT trade-off across the evaluated budgets. \relaxkvg{} provides the complementary matched-target control, isolating the effect of recomputation context under ProphetKV's repair targets. Together, these results show that RelaxKV can make a relatively broad query-relevant KV context available while recomputing only a subset of token states. We use $r=0.15$ for the cross-model comparisons in the main experiments.

% =====================================================
\section{Related Work}
% -----------------------------
\subsection{KV Cache Reuse and Selective Recomputation}

Position-Independent Caching (PIC) enables KV states of independently encoded chunks to be reused across requests with different preceding contexts. Because independently encoded chunks omit cross-chunk interactions, directly composing their KV caches introduces approximation error, motivating online selective repair. CacheBlend, EPIC, KVShare, and ProphetKV address this reusable-cache repair problem with different repair-target selection strategies: EPIC repairs fixed boundary states, CacheBlend and KVShare identify targets using cache or hidden-state deviations, and ProphetKV selects targets according to query relevance \citep{yao2025cacheblend,hu2025epic,yang2025kvshare,wang2026prophetkv}. These methods primarily determine which token states to recompute, while selected targets still attend to their full causal prefixes. RelaxKV operates on the same PIC-based reusable cache without modifying the base model or cache construction. In contrast to prior methods, which select repair targets but recompute them against the full causal prefix, RelaxKV explicitly selects the KV context available during recomputation. This extends selective cache repair from repair-target selection alone to joint allocation over repair targets and recomputation context.

Several related works approach reusable KV caching from broader system or model-design perspectives. Cache-Craft integrates chunk-level KV reuse with cache-variant management, hierarchical storage, and selective recomputation in an end-to-end RAG serving system \citep{agarwal2025cachecraft}. Its scope therefore extends beyond the selective cache-repair problem considered here. Other methods modify the cache construction or model behavior to improve cross-context reuse. KVLink combines positional adjustment with trainable special tokens to recover cross-document interactions \citep{yang2025kvlink}, while KV Packet introduces reusable soft-token boundary adapters learned offline to reduce online recomputation \citep{chen2026kvpacket}. In contrast, RelaxKV leaves the base model and cache construction unchanged and focuses on allocating online recomputation over repair targets and recomputation context.

% -------------------------------------
\subsection{Query-Aware Selection and Sparse Attention}

Several methods use query-related signals to identify important KV states or context positions. SnapKV uses query-side attention to identify salient context positions and compress an already contextualized KV cache \citep{li2024snapkv}. KVzip instead estimates KV importance by reconstructing the original context, producing a compressed cache that is independent of subsequent queries \citep{kim2025kvzip}. These methods reduce the size of an existing contextualized cache, but do not address the missing cross-chunk interactions that arise when independently encoded caches are composed.

Query guidance has also been used to reduce computation during decoding, prefill, and cache repair. ProphetKV, the closest baseline to RelaxKV, uses the user query to select repair targets in a reusable-cache setting \citep{wang2026prophetkv}. Quest uses the current query vector to select KV pages during decoding, while QUOKA selects representative queries and relevant keys to accelerate chunked prefill \citep{tang2024quest,jones2026quoka}. REFORM combines retrieval and recomputation for long contexts by constructing compressed representations across layers, gathering query-relevant input segments, and forwarding the selected inputs to build a compact cache \citep{song2025reform}. These methods demonstrate the value of query-guided selection and sparse computation, but they optimize different stages or cache representations from the selective repair problem considered here.

RelaxKV starts from an existing reusable cache, retains KV entries outside the repair target set, and updates selected entries through online recomputation. More importantly, it separates two decisions that are coupled in prior selective-repair methods: which token states are recomputed and which cached KV states are available during that recomputation.

% --------------------------------
\subsection{Long-Context Evaluation}

LongBench provides a diverse suite of natural long-context tasks, including question answering, multihop reasoning, summarization, and passage retrieval \citep{bai2024longbench}. RULER complements these tasks with controlled synthetic evaluations that probe retrieval and reasoning as context length increases \citep{hsieh2024ruler}. LV-Eval further evaluates long-context understanding in settings with multiple documents and natural distractors \citep{yuan2024lveval}. Together, these benchmarks cover both natural task diversity and controlled length scaling. We use LongBench to evaluate overall task quality, and RULER-MV and LV-Eval to examine whether selective cache repair remains effective as context length increases and irrelevant context becomes more prominent.

% =====================================================
\section{Conclusion}

We introduced RelaxKV, which extends selective cache repair from repair-target selection alone to joint allocation over repair targets and recomputation context. Guided by layer-wise query attention, RelaxKV derives layer-specific repair targets and a shared query-relevant recomputation context. Across four decoder models and multiple long-context benchmarks, RelaxKV achieves competitive quality–latency trade-offs across models and particularly strong performance in long-context settings. Its advantage remains strong as context length increases, with particularly competitive performance on the 16K and 32K RULER-MV and LV-Eval evaluations. Matched-target and context-construction controls further validate the effectiveness of recomputation context selection.

% ====================================================
% \subsection*{AI use statement}

% Generative AI tools were used solely for language polishing, including
% improvements to grammar, wording, clarity, and readability. All AI-assisted
% edits were reviewed by the authors, who take responsibility for the final
% content of the manuscript.

% ====================================================
\bibliography{reference}
\bibliographystyle{iclr2027_conference}

\clearpage
\appendix

% This file supports both standalone compilation and inclusion from the main
% paper. The main paper defines \relaxkvappendixinput before \input.

\providecommand{\relaxkvg}{RelaxKV\smash{\ensuremath{^{\mathrm{G}}}}}
\providecommand{\topk}{\operatorname{TopK}}

% =====================================================
\section{Exact Sparse-Execution Semantics}
\label{app:sparse-recomputation-implementation}
\label{app:exact-method}
% ----------------------------
\subsection{Repair Schedule}

Let $T_c$ denote the number of reusable context tokens and $L$ the number of decoder layers. Query attention produces a layer-specific score $a_{l,i}$ for context token $i$ at layer $l$. Given an anchor ratio $r$, RelaxKV selects
\begin{equation}
    k=\max(1,\operatorname{round}(rT_c))
\end{equation}
repair anchors independently at each layer:
\begin{equation}
    S_l=\operatorname{TopK}_i(a_{l,i},k).
\end{equation}
For each token selected at least once, we define its highest selected layer, the shared anchor union, and the active repair targets as
\begin{align}
    h_i &= \max\{l:i\in S_l\}, \qquad U = \bigcup_l S_l,\\
    R_l &= \{i\in U:h_i\ge l\}.
\end{align}
A token selected at a deeper layer is recomputed through the preceding layers required to reach its highest selected layer $h_i$, and becomes inactive thereafter. Thus, $R_l$ contains exactly the repair targets that execute layer $l$.

For an active repair target at original sequence position $p$, the causal recomputation context is
\begin{equation}
    C_l(p)=U\cap\{j:j\le p\}.
\end{equation}
The same union $U$ is shared across layers, while causal masking determines the subset available to each target. Original global position IDs are retained when applying RoPE, so restricting attention to $C_l(p)$ changes only which KV positions are available during recomputation without renumbering the retained tokens or altering their positional offsets.

\begin{algorithm}[h]
\caption{RelaxKV cache repair}
\label{alg:relaxkv}
\footnotesize
\begin{algorithmic}[1]
\REQUIRE Naive cache $\mathcal{K}^{\mathrm{naive}}$, context tokens $x$, query $q$, anchor ratio $r$
\STATE Run $q$ against $\mathcal{K}^{\mathrm{naive}}$ and obtain layer-wise scores $a_{l,i}$
\STATE $k\leftarrow \max(1,\operatorname{round}(rT_c))$
\FOR{$l=0,\ldots,L-1$}
    \STATE $S_l\leftarrow\operatorname{TopK}_i(a_{l,i},k)$
\ENDFOR
\STATE $U\leftarrow\bigcup_l S_l$
\FOR{each $i\in U$}
    \STATE $h_i\leftarrow\max\{l:i\in S_l\}$
\ENDFOR
\FOR{$l=0,\ldots,L-1$}
    \STATE $R_l\leftarrow\{i\in U:h_i\ge l\}$
\ENDFOR
\STATE Initialize the output cache from $\mathcal{K}^{\mathrm{naive}}$
\STATE Initialize hidden states for positions in $U$ from their input representations
\FOR{$l=0,\ldots,L-1$}
    \STATE Compute Q/K/V for the active repair targets in $R_l$
    \STATE Gather cached K/V states at positions in $U$
    \STATE Replace gathered entries in $R_l$ with freshly recomputed K/V states
    \STATE Apply causal attention so that each target $p\in R_l$ attends only to $C_l(p)$
    \STATE Execute the unchanged output projection, residual connections, normalization, and MLP
    \STATE Update the cache entries corresponding to all targets in $R_l$
\ENDFOR
\RETURN Repaired cache
\end{algorithmic}
\end{algorithm}

% ----------------------------------
\subsection{Recomputation Context and Cache Update}

RelaxKV constructs a compact recomputation context indexed by the shared anchor union $U$. For an active repair target at position $p$ and a context position $j\in C_l(p)$, the KV state used at layer $l$ is
\begin{equation}
 (\widetilde K_{l,j},\widetilde V_{l,j})=
 \begin{cases}
 (K^{\mathrm{new}}_{l,j},V^{\mathrm{new}}_{l,j}),
 & j\in R_l\cap C_l(p),\\
 (K^{\mathrm{naive}}_{l,j},V^{\mathrm{naive}}_{l,j}),
 & j\in C_l(p)\setminus R_l.
 \end{cases}
\end{equation}
Thus, active positions contribute freshly recomputed KV states, while the remaining positions in the recomputation context reuse their cached values. Positions outside $U$ are not directly available during recomputation. After executing the Transformer block, the newly recomputed KV states overwrite the cache entries corresponding to all targets in $R_l$; no additional selection step is required for cache update.

For compact execution, the global positions in $U$ are stored in sorted order and mapped to compact indices once per request. At each layer, cached KV states at these positions are gathered, and entries corresponding to active targets are replaced with freshly recomputed states. Causal masking then restricts each target at position $p$ to
\[
    C_l(p)=U\cap\{j:j\le p\},
\]
while preserving the original global position IDs used by RoPE.

The two selection dimensions affect different parts of the computation. We characterize attention-context exposure by the number of causal attention edges,
\begin{equation}
    \mathcal{E}_{\mathrm{unmasked}}(R,C)
    =
    \sum_l\sum_{p\in R_l}|C_l(p)|,
\end{equation}
while projection and MLP work scale with the number of executed layer-token states,
\begin{equation}
    \operatorname{Cost}_{\mathrm{state}}(R)
    \propto
    \sum_l |R_l|.
\end{equation}
The edge count $\mathcal{E}_{\mathrm{unmasked}}$ is a structural measure rather than an exact FLOP count for our eager grouped-query attention (GQA) implementation. KV heads are expanded to the query-head layout before rectangular attention is executed, and causally masked entries may therefore still incur kernel work. We consequently use measured recomputation latency and TTFT as the primary efficiency metrics. The nominal anchor ratio, realized context-set ratio, and active-target ratio describe different quantities.

% -------------------------------------
\subsection{Controlled Configurations}

Table~\ref{tab:variant-definitions} summarizes the controlled configurations used to isolate the effects of context restriction, cross-layer context sharing, and repair-target construction. Unless otherwise noted, context-construction variants use the same layer-wise anchors $S_l$ and active repair targets $R_l$, so that the primary design variable is the recomputation context.

\begin{table*}[t]
\centering
\caption{Controlled configurations. Every active repair target updates its
corresponding KV entry; configurations vary in active repair targets or
recomputation context columns. Global repair targets $P$ are shared by all
layers.}
\label{tab:variant-definitions}
\small
\setlength{\tabcolsep}{5pt}
\renewcommand{\arraystretch}{1.08}
\resizebox{\textwidth}{!}{%
\begin{tabular}{llll}
\toprule
\textbf{Configuration} & \textbf{Active Repair Targets}
& \textbf{Recomputation Context} & \textbf{Role} \\
\midrule
RelaxKV & $R_l$ & $C_l(p)$ & Main method \\
\relaxkvg{} & Global $P$ & $C_l(p)$
& Matched-target diagnostic control \\
Full-Prefix Control & $R_l$ & Full causal prefix
& Full-prefix quality--cost control \\
Global Sparse & $R_l$ & Global Top-$|U|$
& Global context construction \\
Layerwise Sparse & $R_l$ & Layerwise Top-$|U|$
& Layerwise context construction \\
Global Repair Target Sweep & Global $P_{r_g}$ & $C_l(p)$
& Global target capacity \\
\bottomrule
\end{tabular}}
\end{table*}

% ------------------------------------
\subsection{Full-Prefix Quality--Cost Control}
\label{app:full-prefix-control}

The context-construction ablation in the main paper matches the nominal context-set size before causal truncation, but does not assume equal kernel work. The realized budgets can differ substantially from the nominal anchor ratio. In the Llama-3.1 sweep, a 20\% anchor ratio yields mean context-set ratios of 65.1\% on RULER-MV and 70.4\% on MuSiQue. For Qwen3-14B at a 15\% anchor ratio, the mean context-set and active-target ratios are 64.1\%/44.8\% on MuSiQue, 78.1\%/55.2\% on RULER-MV 32K, and 80.3\%/57.8\% on LV-Eval. These measurements illustrate why the nominal anchor ratio should not be interpreted as either the realized context budget or the fraction of executed layer-token states.

Table~\ref{tab:full-prefix-control} isolates the effect of recomputation context restriction by fixing the active repair targets and varying only the context available to them. The full-prefix control allows each target to attend to its complete causal prefix, whereas RelaxKV uses its restricted recomputation context. Across the four settings, RelaxKV reduces mean recomputation latency by 9.9--23.2\%, while the absolute quality difference remains within 0.45--1.00 points. This experiment therefore characterizes the direct quality--cost trade-off introduced by context restriction and is complementary to the equal-size context-construction ablation in the main paper.

\begin{table*}[t]
\centering
\caption{Full-prefix quality--cost control on Qwen3-14B with a 20\% anchor
ratio. Both configurations share repair anchors, dependency-closed active
targets, global position IDs, and cache updates. Recomp. is mean recomputation
latency per example.}
\label{tab:full-prefix-control}
\small
\setlength{\tabcolsep}{2.4pt}
\resizebox{\textwidth}{!}{%
\begin{tabular}{ll*{8}{c}}
\toprule
\multicolumn{2}{c}{\cellcolor{white}}
& \multicolumn{2}{c}{\textbf{MuSiQue}}
& \multicolumn{2}{c}{\textbf{RULER-MV 32K}}
& \multicolumn{2}{c}{\textbf{LV-Eval 16K}}
& \multicolumn{2}{c}{\textbf{LV-Eval 32K}} \\
\cmidrule(lr){3-4}\cmidrule(lr){5-6}\cmidrule(lr){7-8}\cmidrule(lr){9-10}
\textbf{Variant} & \textbf{Recomputation Context}
& \textbf{F1} $\uparrow$ & \textbf{Recomp. (s)} $\downarrow$
& \textbf{Score} $\uparrow$ & \textbf{Recomp. (s)} $\downarrow$
& \textbf{F1} $\uparrow$ & \textbf{Recomp. (s)} $\downarrow$
& \textbf{F1} $\uparrow$ & \textbf{Recomp. (s)} $\downarrow$ \\
\midrule
Full-Prefix Control & Full causal prefix
& 39.63 & 10.11 & 96.00 & 55.05
& 29.32 & 38.78 & 23.32 & 115.97 \\
\rowcolor{gray!10}
\textbf{RelaxKV}$_{.20}$ & $C_l(p)$
& 39.01 & 8.51 & 95.00 & 42.30
& 28.87 & 32.88 & 23.86 & 104.45 \\
\bottomrule
\end{tabular}}
\end{table*}

% -------------------------------------
\subsection{Cross-Model Quality--TTFT Trade-Off Curves}
\label{app:multi-model-pareto}

\begin{figure}[H]
    \centering
    \includegraphics[width=0.8\textwidth]{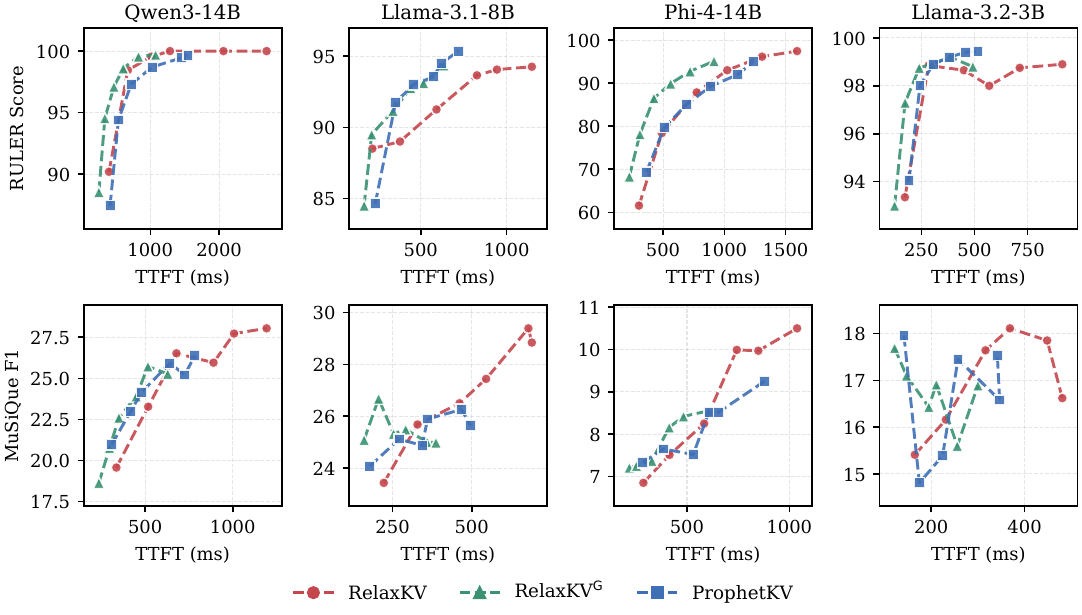}
    \caption{Cross-model quality–TTFT trade-off curves.}
    \label{fig:multi-model-pareto}
\end{figure}

Figure~\ref{fig:multi-model-pareto} extends the Qwen3-14B quality--TTFT analysis in the main paper to all four evaluated models. The top row reports RULER-MV accuracy and the bottom row reports MuSiQue F1, with columns corresponding to Qwen3-14B, Llama-3.1-8B, Phi-4-14B, and Llama-3.2-3B. All panels use the same cache-ready TTFT protocol, and each curve traces the quality--latency trade-off as the nominal repair or anchor ratio varies.

% -----------------------------
\subsection{Chunk Size Sensitivity}
\label{app:chunk-size-ablation}

We vary the reusable chunk size while keeping the model, prompts, tokenization, evaluation examples, and nominal repair ratios fixed. Table~\ref{tab:chunk-size-ablation} reports Qwen3-14B results on MuSiQue without truncation for the chunk sizes evaluated by all methods. Across the evaluated range, all reuse-based methods improve as the chunk size increases, consistent with larger chunks preserving more within-chunk contextual interactions during independent encoding.

\begin{table*}[t]
\centering
\caption{Sensitivity to reusable chunk size on Qwen3-14B MuSiQue. Selective baselines and
\relaxkvg{} use a nominal 20\% repair target ratio; RelaxKV uses a 15\%
anchor ratio. Full Recompute is independent of reusable chunk size and
is repeated as a common dense reference. Bold and underline mark the best
and second-best selective results.}
\label{tab:chunk-size-ablation}
\begingroup
\scriptsize
\setlength{\tabcolsep}{3.4pt}
\renewcommand{\arraystretch}{1.08}
\resizebox{\textwidth}{!}{%
\begin{tabular}{c*{7}{c}>{\columncolor{gray!10}}c}
\toprule
\textbf{Chunk size}
& \shortstack{\textbf{Recompute}}
& \shortstack{\textbf{Reuse}}
& \textbf{CacheBlend}
& \textbf{EPIC}
& \textbf{KVShare}
& \textbf{ProphetKV}
& \relaxkvg{}$_{.20}$
& \textbf{RelaxKV}$_{.15}$ \\
\midrule
64  & \textcolor{gray}{41.89} & \textcolor{gray}{4.45} & 5.54 & 22.54 & 6.40 & \underline{26.77} & 24.94 & \textbf{32.66} \\
128 & \textcolor{gray}{41.89} & \textcolor{gray}{5.27} & 7.33 & 25.49 & 7.81 & \underline{31.75} & 30.08 & \textbf{34.54} \\
256 & \textcolor{gray}{41.89} & \textcolor{gray}{7.81} & 9.64 & 30.49 & 10.46 & \underline{33.97} & 31.62 & \textbf{36.39} \\
512 & \textcolor{gray}{41.89} & \textcolor{gray}{12.62} & 13.64 & 30.19 & 13.41 & 35.57 & \underline{36.28} & \textbf{38.08} \\
1024 & \textcolor{gray}{41.89} & \textcolor{gray}{18.46} & 22.47 & 34.05 & 19.00 & \textbf{39.27} & 37.62 & \underline{38.93} \\
\bottomrule
\end{tabular}}
\endgroup
\end{table*}

% =====================================================
\section{Global Repair Target Capacity Analysis}
\label{app:target-sweep}

\subsection{Controlled Setup}

This diagnostic tests whether a larger global repair target set shared across layers can substitute for RelaxKV's layer-specific active repair targets. We fix the recomputation context $U$ as the union of layer-wise Top-20\% anchors and define the global score and target set as
\begin{equation}
  g_i=\frac{1}{L}\sum_{l=0}^{L-1}a_{l,i},\qquad
  P_{r_g}=\operatorname{TopK}_i\!\left(
  g_i,\min(\lfloor r_gT_c\rfloor,|U|)
  \right).
\end{equation}
The same target set $P_{r_g}$ is recomputed and updated at every layer, while each target attends only to the causal subset of $U$. Targets are ranked over the full context, so the cardinality cap does not require $P_{r_g}\subseteq U$. The \emph{Target-Union} configuration matches only the target count, $|P|=|U|$, rather than setting $P=U$.

\begin{table}[t]
\centering
\caption{Global repair target capacity sweep on Llama-3.1-8B-Instruct. The
shared context is fixed as the union of layerwise Top-20\% anchors. TTFT is in
milliseconds.}
\label{tab:global-target-summary}
\scriptsize
\setlength{\tabcolsep}{3.2pt}
\renewcommand{\arraystretch}{1.05}
\begin{tabular}{crrr}
\toprule
\textbf{Repair Target Ratio}
& \textbf{RULER} $\downarrow$
& \textbf{MuSiQue} $\downarrow$
& \textbf{LB Avg} $\uparrow$ \\
\midrule
0.2 & 361.58 & 267.86 & 57.84 \\
0.3 & 486.23 & 332.68 & 57.24 \\
0.4 & 609.55 & 396.31 & 57.57 \\
0.5 & 753.35 & 458.32 & 57.37 \\
0.6 & 871.97 & 539.33 & 58.06 \\
0.7 & 941.26 & 608.62 & 58.10 \\
Target-Union & 914.47 & 606.73 & 58.10 \\
\bottomrule
\end{tabular}
\end{table}

Increasing the global target ratio substantially raises TTFT while providing limited aggregate quality gains. From 20\% targets to the Target-Union configuration, RULER-MV TTFT increases from 361.58 to 914.47~ms and MuSiQue TTFT from 267.86 to 606.73~ms, corresponding to 2.53$\times$ and 2.27$\times$ increases. Over the same range, the seven-task LongBench average changes only from 57.84 to 58.10 and varies non-monotonically at intermediate ratios. Thus, under a fixed recomputation context, simply increasing the number of globally shared repair targets yields rapidly increasing execution cost with limited aggregate benefit.

Requested target ratios may saturate at the per-example context-set cardinality. Table~\ref{tab:global-target-detail} therefore reports the realized target and context-set ratios together with the per-task scores used in the aggregate. Some examples are already capped at a requested ratio of 0.7, causing the realized ratios to differ across datasets.

\begin{table*}[t]
\centering
\caption{Effective repair-target/context-set ratios and complete LongBench
scores for the global target sweep. Requested target ratios are capped
per example at $|U|$.}
\label{tab:global-target-detail}
\scriptsize
\setlength{\tabcolsep}{3.0pt}
\renewcommand{\arraystretch}{1.04}
\begin{tabular}{lccrrrrrrrr}
\toprule
\textbf{Repair Target}
& \textbf{RULER Ratio} & \textbf{MuSiQue Ratio}
& \textbf{WQA} & \textbf{TQA} & \textbf{HQA} & \textbf{NQA}
& \textbf{MQue} & \textbf{PR-en} & \textbf{PR-zh} & \textbf{Avg} \\
\midrule
0.2 & .200/.651 & .200/.704
& 45.86 & 90.80 & 48.62 & 25.30 & 25.40 & 74.50 & 94.43 & 57.84 \\
0.3 & .300/.651 & .300/.704
& 44.96 & 90.30 & 47.61 & 24.90 & 25.02 & 75.50 & 92.38 & 57.24 \\
0.4 & .400/.651 & .400/.704
& 45.14 & 90.30 & 48.57 & 25.37 & 25.71 & 76.00 & 91.88 & 57.57 \\
0.5 & .500/.651 & .500/.704
& 45.56 & 91.13 & 48.48 & 24.51 & 25.48 & 76.00 & 90.46 & 57.37 \\
0.6 & .600/.651 & .600/.704
& 46.40 & 91.46 & 49.29 & 25.86 & 25.99 & 75.50 & 91.90 & 58.06 \\
0.7 & .651/.651 & .694/.704
& 45.70 & 91.46 & 48.07 & 26.19 & 28.13 & 76.00 & 91.13 & 58.10 \\
Target-Union & .651/.651 & .704/.704
& 45.70 & 91.46 & 48.07 & 26.19 & 28.13 & 76.00 & 91.13 & 58.10 \\
\bottomrule
\end{tabular}
\end{table*}

This diagnostic shows that target count alone does not explain the benefit of a repair schedule that varies across layers. It does not imply that a particular target selector is universally optimal, nor that equal nominal ratios correspond to equal computation. End-to-end quality--latency comparisons remain those reported in the main paper.

% =====================================================
\section{Reproducibility Details}
\label{app:reproducibility}

\subsection{Model, Cache, and Precision}

We evaluate Llama-3.1-8B-Instruct, Qwen3-14B, Phi-4-14B, and Llama-3.2-3B-Instruct in BF16 with batch size one, greedy decoding, and random seed 42. Experiments use PyTorch 2.8.0, CUDA 12.9, and Transformers 4.57.1 on NVIDIA H100 80GB GPUs. All selective methods run on one GPU; only the 16K and 32K full-recomputation references use two-GPU model parallelism.

Reusable contexts are partitioned into 512-token chunks and encoded independently with their final global position IDs. Their KV caches are concatenated to form the naive reusable cache used by all selective methods.

\subsection{Task Definitions and Evaluation Splits}

RULER-MV uses a nominal 8K context in the standard comparison and 4K--32K in the length stress test. LongBench follows the standard task metrics with an 8,192-token input limit and head--tail truncation. The recomputation-context ablation uses untruncated MuSiQue inputs, while LV-Eval is evaluated at 16K and 32K context lengths.

\subsection{Accuracy Protocol}

Within each comparison, all methods use identical prompts, tokenization, model weights, and evaluation examples. The global target-capacity analysis uses Llama-3.1-8B-Instruct. The reported LongBench aggregate is the arithmetic mean of the seven selected task scores.

\subsection{Cache-Ready TTFT Protocol}

TTFT is measured after one untimed warm-up with three timed repeats and one generated token. The reusable naive cache is resident on GPU before timing. For selective methods, timing includes query-attention scoring, target and context selection, KV gathering, recomputation, cache update, query prefill, and first-token decoding. Reusable-cache construction is excluded.

\end{document}